%% file: emnlp2024.tex
\pdfoutput=1

\documentclass[11pt]{article}

\usepackage{ACL2023}

\usepackage{times}
\usepackage{latexsym}

\usepackage[T1]{fontenc}

\usepackage[utf8]{inputenc}

\usepackage{microtype}

\usepackage{inconsolata}

\usepackage{mymathdef}

\newcommand{\FT}[0]{\includegraphics[width=.018\textwidth]{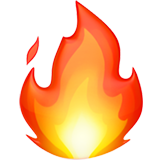}}
\newcommand{\FE}[0]{\includegraphics[width=.018\textwidth]{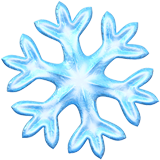}}

\title{RevMUX: Data Multiplexing with Reversible Adapters for Efficient \\LLM Batch Inference}

\author{
Yige Xu${^{1,2}}$, Xu Guo$^1$\thanks{\ \  Corresponding authors.}, Zhiwei Zeng$^1$\footnotemark[1], Chunyan Miao$^{1,2}$ \\
$^1$Joint NTU-UBC Research Centre of Excellence in Active Living for the Elderly \\
$^2$College of Computing and Data Science \\
Nanyang Technological University, Singapore \\
\texttt{\{yige002,xu008\}@e.ntu.edu.sg}, \texttt{\{zhiwei.zeng,ascymiao\}@ntu.edu.sg}
}

\begin{document}
\maketitle
\begin{abstract}
Large language models (LLMs) have brought a great breakthrough to the natural language processing (NLP) community, while leading the challenge of handling concurrent customer queries due to their high throughput demands. Data multiplexing addresses this by merging multiple inputs into a single composite input, allowing more efficient inference through a shared forward pass. However, as distinguishing individuals from a composite input is challenging, conventional methods typically require training the entire backbone, yet still suffer from performance degradation. In this paper, we introduce RevMUX, a parameter-efficient data multiplexing framework that incorporates a reversible design in the multiplexer, which can be reused by the demultiplexer to perform reverse operations and restore individual samples for classification. Extensive experiments on four datasets and three types of LLM backbones demonstrate the effectiveness of RevMUX for enhancing LLM inference efficiency while retaining a satisfactory classification performance. Codes are available at \url{https://github.com/xuyige/RevMUX}.
\end{abstract}

\section{Introduction}

In recent years, Large Language Models (LLMs), such as GPT-3 (175B)~\cite{DBLP:conf/nips/BrownMRSKDNSSAA20}, PaLM (540B)~\cite{chowdhery2023palm}, and GPT-4 (1.7T)~\cite{openai2023gpt4} have emerged as a cornerstone in Natural Language Processing (NLP). The field has witnessed a dramatic increase in model sizes, which, although improving downstream performance, also poses considerable challenges. Inference with these LLMs has become increasingly resource-intensive, often confronting users with capacity limits~\cite{openai2023gpt4}. With the rise of ``language model as a service''~\cite{DBLP:conf/icml/SunSQHQ22}, improving inference efficiency has become a key focus to accommodate these growing model sizes.

To explore efficient inference for LLMs, the community has mainly focused on model-centric or algorithm-centric approaches~\cite{wan2023efficient}. Model-centric approaches, including quantization~\cite{bhandare2019efficient} and knowledge distillation~\cite{hinton2015distilling}, aim to compress LLMs into smaller models while retaining the capabilities of the vanilla models. In contrast, algorithm-centric approaches, such as speculative decoding~\cite{DBLP:conf/icml/LeviathanKM23} and KV-Cache optimization~\cite{DBLP:conf/nips/Zhang00CZC0TRBW23}, aim to reduce latency and memory usage in sequence generation tasks. However, when processing batch queries in a single forward pass, these methods generally result in a significant increase in computational load, e.g., FLOPs, linear with the number of inputs.

\begin{figure*}[!thp]
  \centering
  \subfloat[Mini Batch Processing]{
  \centering
  \includegraphics[width=0.28\textwidth]{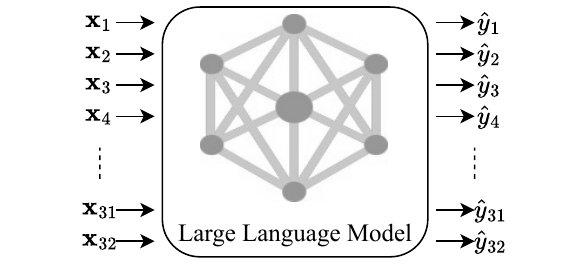}\label{fig:bi-methodology-minibatch}
  }
  \subfloat[DataMUX]{
  \centering
  \includegraphics[width=0.55\textwidth]{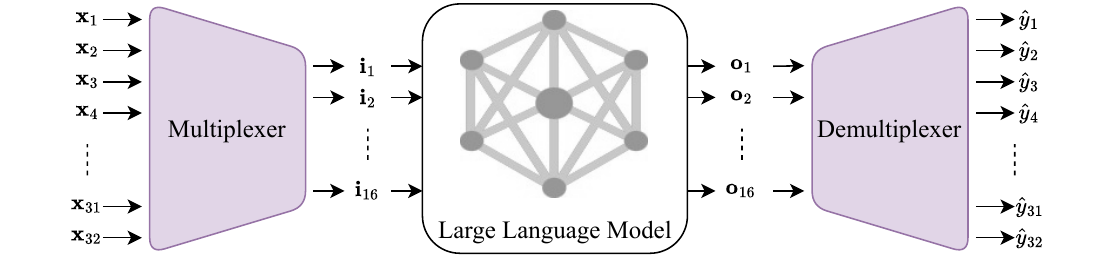}\label{fig:bi-methodology-datamux}
  }\\
  \subfloat[RevMUX]{
  \centering
  \includegraphics[width=0.85\textwidth]{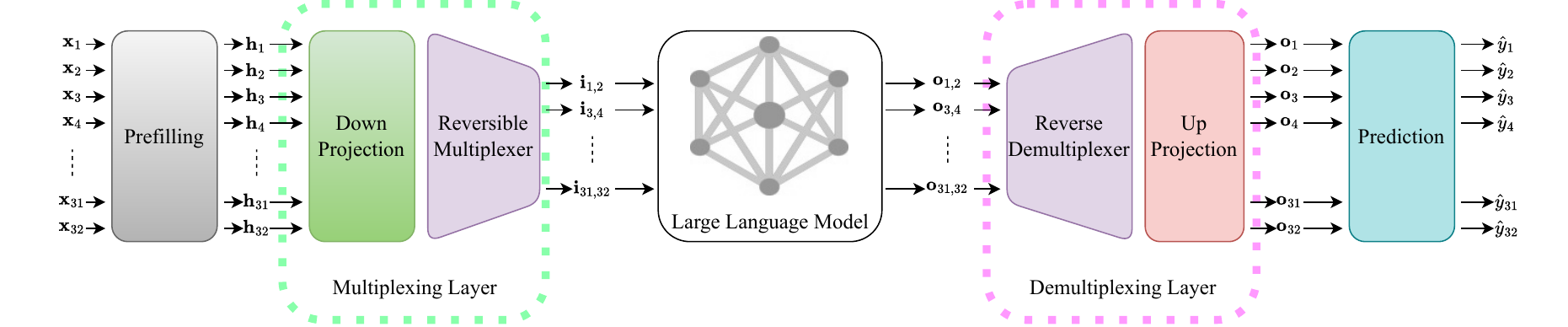}\label{fig:bi-methodology-pipeline}
  }
\caption{Illustration of our proposed RevMUX in comparison to traditional mini-batch processing and DataMUX.}\label{fig:bi-methodology-overview}
\end{figure*}

The Multi-input Multi-output (MIMO) architecture~\cite{DBLP:conf/iccv/RameSC21,DBLP:conf/iclr/HavasiJFLSLDT21,DBLP:conf/nips/MurahariJYN22} has emerged as an effective approach for predicting multiple samples simultaneously in a single forward pass, incurring the computational cost of only a single input. This architecture requires jointly training a multiplexer and a demultiplexer alongside the entire base model: the multiplexer combines multiple inputs into one, and the demultiplexer separates the base model's outputs back into individual ones. Subsequent research~\cite{DBLP:conf/emnlp/MurahariDJSW0N23} applied MIMO-style training to enhance inference with LLMs, exemplified by BERT~\cite{DBLP:conf/naacl/DevlinCLT19}. However, this method not only necessitates training the multiplexer and demultiplexer during pre-training phase but also requires fine-tuning the entire model, including BERT's parameters, thereby limiting its applicability to increasingly larger language models. Moreover, updating the backbone model's parameters necessitates maintaining multiple copies of the backbone model to accommodate dynamic inference budgets, further constraining its scalability.

In this paper, we explore MIMO training on a fixed LLM to improve its inference efficiency without additional pre-training. A major challenge in implementing MIMO for fixed LLMs is to trace and preserve the uniqueness of each input, as the fixed LLMs can struggle to differentiate individual instances within the consolidated inputs, resulting in performance degradation~\cite{DBLP:conf/emnlp/MurahariDJSW0N23}. Inspired by Reversible Neural Networks~\cite{DBLP:conf/nips/GomezRUG17,behrmann2019invertible}, which split the input into two halves for parallel processing and enable reconstruction of lower-layer inputs from upper-layer outputs, we propose reversible adapters to achieve data multiplexing, dubbed RevMUX.
The reversible multiplexer leans to map different samples into distinct feature spaces and the mapping function is shared with the demultiplexer to perform a reverse operation that dis-aggregates the output from the LLM for classification. We train these reversible adapters in a parameter-efficient manner~\cite{DBLP:conf/emnlp/LesterAC21} on downstream tasks and then apply them for batch inference.

Through extensive experiments on four datasets and three types of LLM backbones, we demonstrate the effectiveness of RevMUX in enhancing LLM inference efficiency while maintaining satisfactory classification performance. Notably, our RevMUX method, which freezes the entire backbone LLM, achieves performance comparable to or surpassing that of two fine-tuned baselines on $\mathrm{BERT}_{\mathrm{BASE}}$. We also extended our method to encoder-decoder architectures such as $\mathrm{T5}$ and decoder-only architectures like LLaMA3-8B~\cite{dubey2024llama}. Results on all three architectures across five scales consistently show that the proposed reversible adapters significantly contribute to performance retention during data multiplexing. Moreover, the combined use of reversible adapters in both the multiplexing and demultiplexing processes creates a synergistic effect, amplifying performance benefits beyond those achieved by individual components.

\section{Related Work}

\subsection{Efficient Inference for LLMs}
The majority of recent efforts to enhance LLM inference efficiency have focused on either model-centric or algorithm-centric approaches.

Model-centric methods, also known as model compression techniques, aim to train smaller models that enable efficient inference while retaining the capabilities of the original, larger models. As summarized by \citet{wan2023efficient}, recent model compression techniques for LLMs can be grouped into the following categories: (1) {\it Quantization}, which converts model weights and/or activations of high-precision data types (e.g., float32) into low-precision data types (e.g., int8)~\cite{DBLP:conf/nips/DettmersPHZ23,DBLP:conf/icml/XiaoLSWDH23,shao2024omniquant}; (2) {\it Parameter Pruning}, which removes redundant LLM weights~\cite{DBLP:conf/nips/MaFW23,DBLP:conf/icml/FrantarA23}; and (3) {\it Knowledge Distillation}, which involves training a small student model to mimic the behavior of a large teacher model~\cite{gu2024minillm,DBLP:conf/coling/LiuZKKJS024}.

In contrast, algorithm-centric methods focus on optimizing the inference process through the design of time- or memory-efficient algorithms. For example, speculative decoding supports parallel token computation for autoregressive language models, thereby speeding up the decoding stage~\cite{DBLP:conf/icml/LeviathanKM23,DBLP:conf/acl/SantilliSPMMMR23}. Additionally, KV-Cache optimization, which reuses cached KV pairs, can reduce the computational cost of decoding~\cite{DBLP:conf/nips/Zhang00CZC0TRBW23,ge2024model}.

The above methods either compress the models or optimize the inference process but do not leverage data-specific strategies. When applied to batch queries in a single forward pass, they typically result in a significantly increased computational load, often proportional to the number of inputs. In contrast, our approach enhances inference efficiency through a data-centric strategy. We focus on data multiplexing techniques instead of modifying models or algorithms, allowing the model to perform batch inference with significantly reduced computational costs.

\subsection{Multi-Input Multi-Output Training}

To reduce both training and inference costs in ensemble models, the concept of Multi-Input Multi-Output (MIMO)~\cite{DBLP:conf/iclr/HavasiJFLSLDT21} has been introduced. MIMO enables the training of multiple independent subnetworks within a single network, thereby enhancing prediction robustness. This mechanism allows for multiple output predictions through a single forward pass, effectively simulating the ensemble process while conserving computational resources~\cite{DBLP:conf/iccv/RameSC21,DBLP:conf/cvpr/SunRMTC22,DBLP:journals/pr/SunMHTC24}.
Although previous MIMO works primarily focus on enhancing ensemble efficiency, their findings crucially substantiate the ability of deep neural networks to process multiple inputs in a single forward pass, laying the ground for subsequent works on data multiplexing.

Recent works have lent MIMO-style training to improve the batch inference efficiency of LLMs. \citet{DBLP:conf/nips/MurahariJYN22} propose a data mixer to amalgamate multiple inputs and a corresponding demixer to disaggregate the combined output into individual ones. Specifically, within a batch of $N\times M$ instances, a multiplexing layer consolidates these $N\times M$ representations into $M$ consolidated representations. Subsequently, the demultiplexing layer interprets these $M$ outputs to generate predictions for the entire set of $N\times M$ instances. This approach, embodied in MIMONets~\cite{DBLP:conf/nips/MenetHKBSR23}, incorporates a distinctive key mechanism that serves not only to bind the inputs together but also facilitates their separation. Building upon this concept, MUX-PLMs~\cite{DBLP:conf/emnlp/MurahariDJSW0N23} have advanced the field by pre-training language models that leverage a contextual multiplexer coupled with an RSA demultiplexer, marking a significant step forward in the efficient inference on PLMs.

However, existing MIMO-style frameworks for LLM batch inference typically require end-to-end training, where the base model is trained alongside the data mixer and demixer. This would become impractical for large-scale language models due to their substantial size and complexity. Consequently, this paper focuses on scenarios where the backbone model is already trained and fixed, exploring strategies for effective data multiplexing without any additional pre-training.

\section{Methodology}

\subsection{Overview of RevMUX}
\label{sec:bi-method-overview}

Given an input instance $x$ and a LLM $f(\cdot)$, most existing LLM applications can be summarized as:
\begin{align}
  \hat{y}=f(x),
  \label{eq:def-ml-inference}
\end{align}
where $\hat{y}$ is the prediction. During the inference stage, take Figure~\ref{fig:bi-methodology-overview} as an example, traditional mini-batch processing extends input vector into tensors to improve the throughput. DataMUX~\cite{DBLP:conf/nips/MurahariJYN22} introduce a multiplexer to combine 32 input samples into 16 vectors and a demultiplexer to decompose 16 outputs to 32 labels, which saves the computational load because the LLM only infer ``16 samples''.
Due to the challenge of fixing backbone LLM, the main difference between our proposed RevMUX and DataMUX are two folds:

{\bf Prefilling}:
The mixture of input samples may lead to a distribution shift~\cite{navarro2024data}, which makes the gap on latent representation space between the backbone language model and the multiplexed input samples. Hereby the decision to tine-tune the backbone language model versus not fine-tuning it represents two distinct methodological approaches. Fine-tuning adapts the backbone language model to new tasks by mixing multiple input samples and learning their ralational representations. In contrast, not fine-tuning corresponds to learn how to mix the input samples by bridging the gap between different representation space. To tackle this problem, we use prefilling for transforming the feature space, to ensure the feature space becoming more similar to the feature space seen during pre-training.

{\bf Reversible Multiplex Adapter and Reverse Demultiplex Adapter}:
While combining feature vectors of multiple samples into a single vector can reduce the computational load, such processing can result in significant information loss and model confusion. To preserve the distinction between different samples, it is essential to incorporate a traceable module. Such as module should effectively revert and separate the combine features, ensuring that each sample's unique characteristics are retained for accurate classification. Drawing inspiration from Reversible Neural Networks~\cite{DBLP:conf/nips/GomezRUG17,behrmann2019invertible}, which divide the input into two halves to facilitate the reconstruction of lower layer activations from the upper layer outputs, we introduce reversible adapters to mix and demix different samples within a batch. These adapters are trained in a parameter-efficient manner on downstream tasks and are then employed for batch inference. In Section~\ref{sec:bi-methodology-revmux-pipeline}, we introduce our RevMUX pipeline when $N=2$. Details of multiplexing layer and demultiplexing layer when $N>2$ can be found in Appendix~\ref{sec:bi-appendix-n-gt-2}.

\subsection{The RevMUX Pipeline}
\label{sec:bi-methodology-revmux-pipeline}

\subsubsection{Task-specific Backbone}
\label{sec:bi-methodology-ts-backbone}

Our work aims to address the problem where users, having a large language model already in place for their target task, seek to accelerate inference. In the initial step, it is crucial to have a backbone model capable of addressing the target task. In this paper, we selected T5~\cite{DBLP:journals/jmlr/RaffelSRLNMZLL20} as the backbone to experimentally validate the effectiveness of our approach. Additionally, for comparison purposes, we also utilized BERT~\cite{DBLP:conf/naacl/DevlinCLT19} as the backbone in our comparative experiments.
For T5, we fix the language model and use prompt tuning~\cite{DBLP:conf/acl/LiuJFTDY022} to train a soft prompt, which simulates the scenarios that we cannot train a task-specific large language model. For BERT, we simply add a classfier and fine-tune the BERT to learn the task-specific backbone. For LLaMA3-8B~\cite{dubey2024llama}, we are not able to train the backbone, hereby we assume that the backbone is well trained and can be directly transferred to our classification tasks.

\subsubsection{Prefilling}

As shown in Section~\ref{sec:bi-method-overview}, the first step of RevMUX pipeline is prefilling. In this step, we convert the input instances $x_1,x_2,\cdots,x_N$ to dense representations, ensuring the feature space becoming more similar to the feature space seen during the backbone pre-training:
\begin{align}
  \bh_k^l=\mathrm{Encoder}_{0:l}(\bX_k),
\end{align}
where $l$ indicates that we use the first $l$ encoder layers of the pre-trained language model for prefilling. For BERT and LLaMA, $\bX_k=x_k$. For T5, $\bX_k=\mathrm{concat}[\bp_0;x_k]$ where $\bp_0$ is a pre-trained task-specific soft prompt.

\subsubsection{Multiplexing Layer}
\label{sec:bi-methodology-grouping}

With prefilling, we obtain $N$ representations for $N$ instances. Then we have a multiplexing layer $g: \mbbR^{N\times d}\to \mbbR^d$ to mix instances together, where $d$ is the dimension of the backbone language model. As shown in Figure~\ref{fig:bi-methodology-pipeline}, our multiplexing layer includes down projection and reversible multiplexer.

\paragraph{Down Projection} Since the dimension of backbone language model is $d$ and the representations of instances are under the space of $\mbbR^{N\times d}$, we firstly employ a linear layer $f_{\mathrm{down}}:\mbbR^d \to \mbbR^{\frac{d}{N}}$ as the down projection function:
\begin{align}
  \bi_k^l=f_{\mathrm{down}}(\bh_k^l).
\end{align}

\paragraph{Reversible Multiplexer}
Motivated by Reversible Neural Network~\cite{DBLP:conf/nips/GomezRUG17} that the layers' activations can be reconstructed from the next layers, we employ a reversible multiplexer to combine the multiple input instances, which the demultiplexing layer can reconstruct.

\begin{figure*}
  \centering
  \includegraphics[width=0.9\textwidth]{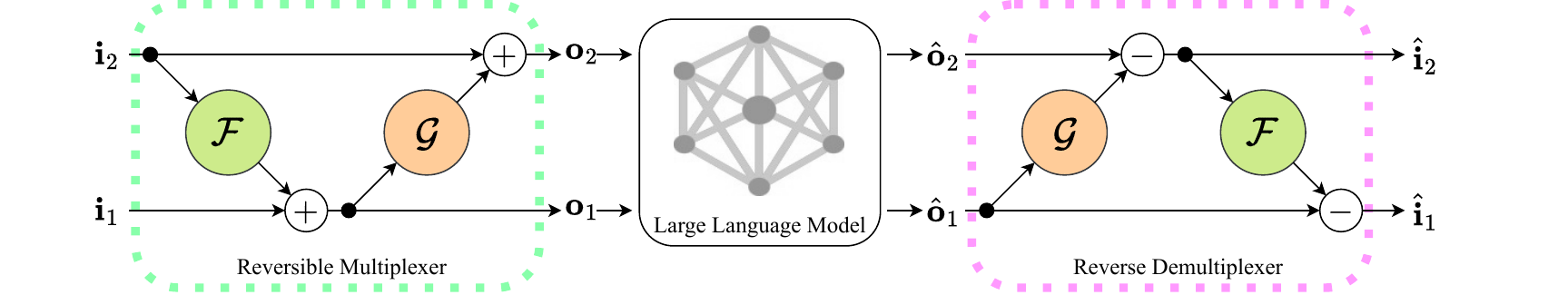}

  \caption{Illustration of the reversible multiplexer and reverse demultiplexer when $N=2$.}
  \label{fig:bi-method-invertible}
\end{figure*}

As illustrated in Figure~\ref{fig:bi-method-invertible} when $N=2$, we have:
\begin{align}
  \label{eq:bi-methodology-ig}
  \bo_1^l&=\bi_1^l+\cF(\bi_2^l),\\\nonumber
  \bo_2^l&=\bi_2^l+\cG(\bo_1^l),\\\nonumber
  \bo^l&=\mathrm{concat}[\bo_1^l,\bo_2^l],
\end{align}
where $\cF(\cdot)$ and $\cG(\cdot)$ are learnable 2-layer MLP.

\subsubsection{Language Modeling with Batched Instances}

With the multiplexing layer, we obtain $\bo_l$ that contains the representation of all batched instances. After that, we pass the batched representation throughout the backbone language model:
\begin{align}
  \hat{\bo}=\mathrm{Decoder}\Big(\mathrm{Encoder}_{l+1:L}(\bo^l)\Big),
  \label{eq:bi-lm-with-bi}
\end{align}
where $L$ is the number of encoder layers. Specially, for encoder-only backbone such as BERT, $\mathrm{Decoder}(x)=x$.

\subsubsection{Demultiplexing Layer}

From Eq~\eqref{eq:bi-lm-with-bi}, we obtain the outputs of the language model. To demix the outputs, we have a demultiplexing layer $h: \mbbR^d \to \mbbR^{N\times d}$. Similar to the multiplexing layer mentioned in Section~\ref{sec:bi-methodology-grouping}, our demultiplexing layer includes reverse demultiplexer and up projection.

\paragraph{Reverse Demultiplexer}

Given the necessity to decompose the language model's output to restore the outputs of the original samples, we employ a reversible multiplexer during the input content assembly process. Therefore, we use the reverse demultiplexer to decompose the output. Take $N=2$ as an example, we have:
\begin{align}
  \label{eq:bi-methodology-dg}
  [\hat{\bo}_1,\hat{\bo}_2]&=\hat{\bo},\\\nonumber
  \hat{\bi}_2&=\hat{\bo}_2-\cG(\hat{\bo}_1),\\\nonumber
  \hat{\bi}_1&=\hat{\bo}_1-\cF(\hat{\bi}_2),
\end{align}
where $\cF(\cdot )$ and $\cG(\cdot )$ share the same parameters with that in Eq~\eqref{eq:bi-methodology-ig}.

\paragraph{Up Projection}
Considering that we obtain a $\frac{d}{N}$-dimensional sample representation through the reverse demultiplexer, it is necessary to employ an up projection to restore this representation to the original $d$-dimensional space for further processing by the backbone language model:
\begin{align}
  \hat{\bh}_k=f_{\mathrm{up}}(\hat{\bi}_k),
  \label{eq:bi-demultiplex-up-projection}
\end{align}
where $k\in \{1,2,\cdots,N\}$ that indicates the output of the $k$-th instance, $f_{\mathrm{up}}:\mbbR^{\frac{d}{N}} \to \mbbR^d$ is a linear layer for up projection.

\subsubsection{Prediction}

The last step of RevMUX is prediction, which converts the output from the demultiplexing layer to target labels. For BERT, the encoder-only backbone, we reuse the trained task-specific classifier layer:
\begin{align}
  \hat{y}_k=\mathrm{softmax}(W_c\hat{\bh}_k),
\end{align}
where $W_c$ is the task-specific parameter matrix trained in Section~\ref{sec:bi-methodology-ts-backbone}.

For T5 and LLaMA, we reuse the language model head to decode the target token and then use a verbalizer $\cV$ to convert the target token to the target label:
\begin{align}
  \hat{y}_k=\cV(\mathrm{LM\_Head}(\hat{\bh}_k)).
\end{align}

In summary, the overall framework can be abstracted as:
\begin{align}
  \hat{y}_1,\hat{y}_2,\cdots,\hat{y}_N=h\Big(f\big(g(x_1,x_2,\cdots,x_N)\big) \Big),
  \label{eq:def-batch-inference}
\end{align}
where $N$ indicates the number of mixed samples, $f(\cdot)$ indicates the backbone LLM, $g: \mbbR^{N\times d}\to \mbbR^d$ indicates the multiplexing layer, and $h: \mbbR^d \to \mbbR^{N\times d}$ indicates the demultiplexing layer. It is notably that traditional mini-batch processing of Eq~\eqref{eq:def-ml-inference} is a special case of Eq~\eqref{eq:def-batch-inference} under the condition of $N=1$ and $g(x)=h(x)=x$.

\subsection{Training Objectives}

In this subsection, we will briefly introduce our training objectives.

\paragraph{Gold Label}
The first objective function is a task-specific loss function from gold label. In this paper, we use cross-entropy loss as:
\begin{align}
  \cL_{\mathrm{ce}}=- \frac{1}{N} \sum_{i=1}^{N} \sum_{c=1}^{C} y_{i,c} \log(\hat{y}_{i,c}),
\end{align}
where $C$ is the number of labels.

\paragraph{InfoNCE}
On the other hand, considering the need to reconstruct the outputs of the original samples, the second objective function must impose constraints to ensure that the results obtained from multiplexed batch inference closely match those from the original one-by-one forward propagation. This ensures that the remaining components of the backbone language model function correctly. To address this problem, we employ Information Noise-Contrastive Estimation (InfoNCE)~\cite{oord2018representation} as the second objective function. InfoNCE is a loss function used in contrastive learning to maximize the mutual information between positive pairs of samples while minimizing it between negative pairs.

During the training stage, we compute the output representation by twice: one from the multiplexed batch inference, and the other from the original one-by-one forward pass. Within the same batch, we treat the output pairs corresponding to the same sample as positive examples and the remaining output pairs as negative examples. Hereby we compute the loss by:
\begin{align}
  &\cL_{\mathrm{info}}=\sum_{k=1}^{N}{\mathrm{InfoNCE}(\hat{\bh}_k, \bh_k)},\\\nonumber
  =&\sum_{k=1}^{N}-\mathbb{E}[\log \frac{\exp(\hat{\bh}_k\cdot\bh_k)}{\exp(\hat{\bh}_k\cdot\bh_k)+\sum_{j\ne k}^{N}\exp(\hat{\bh}_k\cdot\bh_j)}]
\end{align}
where $\bh_k=\mathrm{LLM}(\bX_k)$ is the output of one-by-one forward pass and $\hat{\bh}_k$ is defined in Eq~\eqref{eq:bi-demultiplex-up-projection}.

Thus, the overall objective is:
\begin{align}
  \label{eq:bi-loss-lambda}
  \cL=\cL_{\mathrm{ce}}+\lambda \cL_{\mathrm{info}},
\end{align}
where $\lambda$ is the weight to control the importance of cross-entropy loss and the InfoNCE loss.

\section{Experiments}

\subsection{Datasets and Evaluation Settings}

We conduct experiments on four datasets across three tasks. For the sentiment classification task, we use SST-2~\cite{socher2013recursive}. For the paraphrase detection task, we use MRPC~\cite{DBLP:conf/acl-iwp/DolanB05}. For the natural language inference task, we use RTE~\cite{DBLP:conf/iclr/WangSMHLB19} and QNLI~\cite{DBLP:conf/iclr/WangSMHLB19}. For fair comparisons with baseline methods, we use ${\mathrm{BERT}}_{\mathrm{BASE}}$ as the backbone. We further examined RevMUX on ${\mathrm{T5}}$ across three different scales.

To better simulate real-world randomness, we conduct 10 tests for each model. In each test, we begin by dividing the samples into $N$ distinct subsets. From each subset, we randomly select a sample to create a batch. This batch is then processed by the model. Given these testing parameters, it is possible for the same sample to yield varying prediction results across different tests. To account for this variability, we calculate the average of these multiple tests to serve as our final evaluation metric. This averaged metric is intended to represent the expected accuracy of the overall sample set in real-world inference scenarios. More details can be found in Appendix~\ref{sec:bi-appendix-testing-rounds}.

\subsection{Baselines}

We consider the following baselines:

\paragraph{DataMUX}~\cite{DBLP:conf/nips/MurahariJYN22}. A MIMO-style learning framework that trains a multiplexing layer to combine a group of $N$ data samples into a single representation and a demultiplexing layer to separate this into $N$ representations for classification. The two layers are typically linear layers trained together with the entire base model.

\paragraph{MUX-PLM}~\cite{DBLP:conf/emnlp/MurahariDJSW0N23}. Also a MIMO-style learning framework, particularly designed for enhancing the throughput for a pretrained LLM. MUX-PLM requires training the multiplexing and demultiplexing layers during the pre-training stage for ${\mathrm{BERT}}_{\mathrm{BASE}}$ to learn the new task of ``combining multiple input samples''. In the experiment section, we use $\mathrm{MUX}$-${\mathrm{BERT}}_{\mathrm{BASE}}$ to indicate this baseline for clarity.

\paragraph{Vanilla Adapters}
It directly applies a vanilla three-layer Multilayer Perception (MLP) for multiplexing and demultiplexing respectively, akin to DataMUX. This baseline examines the effectiveness of the reversible design in RevMUX.

\paragraph{Only Multiplexer Reversible}
It keeps the reversible multiplexer of RevMUX but replaces its demultiplexer with a vanilla three-layer MLP. This baseline empirically examines whether the demultiplexer of RevMUX can restore individual inputs.

\section{Results and Analysis}

\begin{table*}[t!]
\setlength{\tabcolsep}{2pt}
    \centering\small
    \begin{tabular}{c l| c c c c c c c c c c}
    \toprule
    ~ & Model & $N$ & $\nearrow$ & Tuned & Params & SST-2 & MRPC & RTE & QNLI & Avg. Score \\
    \midrule
    \multirow{2}{*}{Backbones} & $\mathrm{BERT}_{\mathrm{BASE}}$~\cite{DBLP:conf/naacl/DevlinCLT19} & 1 & - & \FT & 110M & 92.20 & 87.01 & 62.96 & 90.55 & 83.18  \\
    ~ & $\mathrm{MUX}$-${ \mathrm{BERT}}_{\mathrm{BASE}}$~\cite{DBLP:conf/emnlp/MurahariDJSW0N23} & 1 & 100\% & \FT & 112M & 91.74 & 87.75 & 63.18 & 90.54 & 83.30 \\
    \midrule
    \midrule
    \multirow{2}{*}{Baselines} & $\mathrm{DataMUX}$~\cite{DBLP:conf/nips/MurahariJYN22} & 2 & 180\% & \FT & 166M & 90.50 & 85.05 & \underline{60.87} & \underline{88.39} & 81.20 \\
    ~ & $\mathrm{MUX}$-${ \mathrm{BERT}}_{\mathrm{BASE}}$~\cite{DBLP:conf/emnlp/MurahariDJSW0N23} & 2 & 201\% & \FT & 112M & 90.62 & 83.77 & 58.19 & 88.17 & 80.19 \\
    \midrule
    \multirow{4}{*}{Ours} & Vanilla Adapters & 2 & 156\% & \FE & 16.53M & 90.42 & 84.78 & 60.06 & 88.19 & 80.86 \\
    ~ & Only Multiplexer Reversible & 2 & 161\% & \FE & 20.07M & 90.65 & 84.60 & 60.41 & 88.14 & 80.95 \\
    ~ & RevMUX (\FE) & 2 & 154\% & \FE & 9.45M & \underline{90.85} & \underline{85.06} & 60.72 & 88.25 & \underline{81.22} \\
    ~ & RevMUX (\FT) & 2 & 154\% & \FT & 120M & {\bf 91.21} & {\bf 85.78} & {\bf 61.41} & {\bf 88.72} & {\bf 81.78} \\
    \bottomrule
\end{tabular}
\caption{Model comparison using $\mathrm{BERT}_{\mathrm{BASE}}$ as backbone model. ``\FT'' indicates fine-tune the $\mathrm{BERT}$, ``\FE'' indicates freeze the $\mathrm{BERT}$ as feature extraction only. ``Params'' is the number of learnable parameters. Best results in {\bf bold} and the second-best in \underline{underline}. Inference speedups ($\nearrow$) are reported against the $N=1$ setting.
\label{table:bi-result-comparison}}
\end{table*}

\subsection{Comparison with Baselines}

For a fair comparison with previous methods that involve fine-tuning the backbone model, we first experiment on $\mathrm{BERT}_{\mathrm{BASE}}$ (110M) and also report the fine-tuned results, as shown in Table~\ref{table:bi-result-comparison}.

\noindent (1) \textbf{RevMUX retains performance stably}:
Overall, RevMUX (\FT) consistently outperforms $\mathrm{MUX}$-${\mathrm{BERT}}_{\mathrm{BASE}}$ (\FT) ($p=0.015$) and $\mathrm{DataMUX}$ (\FT) ($p=0.02$) across all four datasets. More importantly, our RevMUX (\FE), which freezes the entire backbone LLM, achieves comparable or superior performance to the two fine-tuned baselines, albeit with some sacrifice in inference efficiency. Notably, RevMUX (\FE) outperforms $\mathrm{MUX}$-${\mathrm{BERT}}_{\mathrm{BASE}}$ (\FT) which requires an additional pre-training stage ($p=0.166$). These results highlight RevMUX's advantage in retaining classification performance during data multiplexing.

\noindent (2) \textbf{No fine-tune scenario is significantly more challenging}: Comparing RevMUX (\FE) with RevMUX (\FT), it is clear that finetuning the backbone LLM significantly enhances performance across all the datasets ($p<0.01$). As the task is very challenging, fine-tuning LLMs proves to bring limited gains.

\noindent (3) \textbf{Components in RevMUX are effective}:
Moreover, RevMUX (\FE) surpasses Vanilla Adapters (\FE), highlighting the effectiveness of reversible design in boosting accuracy. Vanilla Adapters (\FE) performed similarly to Only Multiplexer Reversible (\FE), suggesting that the reversible multiplexer alone offers limited benefits. The effectiveness of RevMUX (\FE) lies in the synergy between the reversible multiplexer and reverse demultiplexer, as shown by comparing RevMUX (\FE) against Only Multiplexer Reversible (\FE).

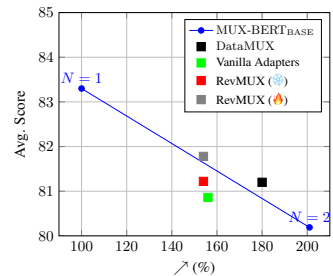
\begin{figure}[!h]
    \centering
    \begin{tikzpicture}[scale=0.52]
        \begin{axis}[
            xlabel={$\nearrow$ (\%)},
            ylabel={Avg. Score},
            legend pos=north east,
            legend cell align={left},
            legend style={font=\small},
            grid=both,
            xmin=90, xmax=210,
            ymin=80, ymax=85
        ]

        \addplot[
            color=blue,
            mark=*,
            nodes near coords,
            point meta=explicit symbolic
        ] coordinates {
            (100, 83.30) [$N=1$]
            (201, 80.19) [$N=2$]
        };
        \addlegendentry{$\mathrm{MUX}$-${ \mathrm{BERT}}_{\mathrm{BASE}}$}

        \addplot[
            only marks,
            mark=square*,
            mark size=3pt
        ] coordinates {
            (180, 81.20) []
        };
        \addlegendentry{$\mathrm{DataMUX}$}

        \addplot[
            only marks,
            mark=square*,
            color=green,
            mark size=3pt
        ] coordinates {
            (156, 80.86) []
        };
        \addlegendentry{Vanilla Adapters}

        \addplot[
            only marks,
            mark=square*,
            color=red,
            mark size=3pt
        ] coordinates {
            (154, 81.22) []
        };
        \addlegendentry{RevMUX (\FE)}

        \addplot[
            only marks,
            mark=square*,
            color=gray,
            mark size=3pt
        ] coordinates {
            (154, 81.78) []
        };
        \addlegendentry{RevMUX (\FT)}

        \end{axis}
    \end{tikzpicture}
    \caption{Trade-off between inference efficiency and model accuracy.}
    \label{fig:bi-experiment-trade-off}
\end{figure}

\noindent (4) \textbf{The trade-off between efficiency and accuracy}: Apart from accuracy, we also measure the total number of FLOPS required for each model to infer all four validation sets. For a fair comparison, we fix the batch size as 32 and the sequence length as 128. We compute the speedups ($\nearrow$) against the baseline $\mathrm{MUX}$-${\mathrm{BERT}}_{\mathrm{BASE}} (N=1)$, reported in the column $\nearrow$ in Table~\ref{table:bi-result-comparison}. We observe that $\mathrm{MUX}$-${\mathrm{BERT}}_{\mathrm{BASE}} (N=2)$, halved the FLOPs consumption, achieving a speedup of 201\% while our RevMUX achieved speedups ranging between 154\% and 161\%, demonstrating a trade-off between model accuracy and efficiency. We present the results in Figure \ref{fig:bi-experiment-trade-off}, where the blue line indicates that the baseline model accuracy decreases as efficiency increases. For a given efficiency target, RevMUX (\FT) and DataMUX (\FT) are clearly above the blue line but RevMUX (\FT) results in higher accuracy, indicating that reversibility can help preserve the classification performance.

\begin{table*}[t!]
\setlength{\tabcolsep}{6pt}
    \centering\small
    \begin{tabular}{l| l c c c c c c c c c c}
    \toprule
    Backbone & Model & $N$ & Tuned & $\nearrow$ & SST-2 & MRPC & RTE & QNLI & Avg. Score \\
    \midrule
    \multirow{4}{*}{$\mathrm{T5}_{\mathrm{Small}}$} & Task-specific Backbone & 1 & \FT & 100\%
    & 90.34 & 84.31 & 64.62 & 89.34 & 82.15 \\
    \cline{2-10}
    ~ & Vanilla Adapters & 2 & \FE & 138\%
    & 89.00 & 81.72 & 57.22 & 85.36 & 78.33 \\
    ~ & Only Multiplexer Reversible & 2 & \FE & 146\%
    & 89.04 & 82.30 & 57.51 & 85.44 & 78.57 \\
    ~ & RevMUX & 2 & \FE & 145\%
    & {\bf 89.14} & {\bf 82.45} & {\bf 60.22} & {\bf 85.63} & {\bf 79.36} \\
    \midrule
    \midrule
    \multirow{4}{*}{$\mathrm{T5}_{\mathrm{Base}}$} & Task-specific Backbone & 1 & \FT & 100\%
    & 94.56 & 87.50 & 82.31 & 92.93 & 89.33 \\
    \cline{2-10}
    ~ & Vanilla Adapters & 2 & \FE & 140\%
    & 92.36 & 82.94 & 63.28 & 87.58 & 81.54 \\
    ~ & Only Multiplexer Reversible & 2 & \FE & 144\%
    & 92.54 & 83.19 & 64.01 & 88.14 & 81.98 \\
    ~ & RevMUX & 2 & \FE & 144\%
    & {\bf 92.70} & {\bf 83.80} & {\bf 64.73} & {\bf 88.65} & {\bf 82.47} \\
    \midrule
    \midrule
    \multirow{4}{*}{$\mathrm{T5}_{\mathrm{Large}}$} & Task-specific Backbone & 1 & \FT & 100\%
    & 95.96 & 90.44 & 87.36 & 93.94 & 91.93 \\
    \cline{2-10}
    ~ & Vanilla Adapters & 2 & \FE & 141\%
    & 92.58 & 83.16 & 64.22 & 88.42 & 82.10 \\
    ~ & Only Multiplexer Reversible & 2 & \FE & 143\%
    & 92.67 & 83.46 & 64.43 & 88.56 & 82.28 \\
    ~ & RevMUX & 2 & \FE & 143\%
    & {\bf 92.81} & {\bf 83.86} & {\bf 65.01} & {\bf 88.89} & {\bf 82.64} \\
    \bottomrule
\end{tabular}
\caption{$\mathrm{T5}$ results on the four datasets of GLUE benchmark. ``\FT'' indicates parameter-efficient fine-tune the $\mathrm{T5}$, ``\FE'' indicates freeze the whole backbone while training the adapters.\label{table:bi-exp-glue-main}}
\end{table*}

\subsection{Scalability Tests on Larger Models}

\subsubsection{Encoder-Decoder Architecture}
In this section, we focus on evaluating our proposed parameter-efficient RevMUX (\FE) on larger language models, specifically on $\mathrm{T5}$. We conduct experiments on $\mathrm{T5}$ with three model sizes: $\mathrm{T5}_{\mathrm{Small}}$ (60M), $\mathrm{T5}_{\mathrm{Base}}$ (220M), and $\mathrm{T5}_{\mathrm{Large}}$ (770M). For each task, we use prompt tuning~\cite{DBLP:conf/emnlp/LesterAC21} to adapt each model to the task domain in advance and then train RevMUX for inference acceleration. The results are presented in Table~\ref{table:bi-exp-glue-main}, and we highlight the following observations:

\noindent (1) {\bf RevMUX retains performance stably while improving efficiency}: We use the result of fine-tuning the entire backbone on each dataset and inference with a single input ($N=1$) as the reference point. When inference with two inputs simultaneously ($N=2$), RevMUX achieves about $45\%$ speedups across all scales, while at the same time, maintaining a satisfactory classification accuracy. This observation holds across the datasets and model scales, demonstrating the generalizability and scalability of RevMUX.

\noindent (2) {\bf Both the reversible multiplexer and reverse demultiplexer are effective}: The findings with the batch inference results ($N=2$) are consistent with those on $\mathrm{BERT}_{\mathrm{BASE}}$. The comparisons between RevMUX and Vanilla Adapters provide strong empirical evidence for the effectiveness of the reversible design in retaining performance. Furthermore, RevMUX consistently surpasses the Only Multiplexer Reversible method in all scenarios, highlighting the synergistic effect of the reversible multiplexer and the reverse demultiplexer.

\noindent (3) {\bf The efficiency-performance trade-off is more pronounced for larger backbones}: The efficiency-performance trade-off is a well-known challenge in the community. Our experiments across various backbone sizes provide empirical evidence that, with a data multiplexing approach, larger backbones experience greater performance compromises in exchange for improved efficiency. Apart from QNLI, the amount of performance degradation on the other datasets follows the trend: $\mathrm{T5}_{\mathrm{Large}} > \mathrm{T5}_{\mathrm{Base}} > \mathrm{T5}_{\mathrm{Small}}$.

\subsubsection{Decoder-Only Architecture}
We now shift our focus to evaluating RevMUX (\FE) on larger decoder-only language models, specifically LLaMA3-8B. Unlike our previous study with T5, we do not pre-adapt LLaMA3 using prompt tuning. Instead, we focus on zero-shot inference, which is commonly employed in billion-scale LLMs. In this study, we assess how RevMUX enhances inference efficiency in a zero-shot context. For each task, we curate a manual prompt and directly train RevMUX on top of LLaMA3 for inference. Additional details can be found in Appendix~\ref{sec:bi-appendix-scaling-llama}. Based on the results presented in Table~\ref{table:bi-exp-glue-llama}, we draw the following key observations:

\noindent (1) {\bf RevMUX is scalable to billion-scale decoder-only LLMs}: Similar to the outcomes observed with $\mathrm{BERT}_{\mathrm{BASE}}$ and three $\mathrm{T5}$ models, both the reversible multiplexer and the reverse demultiplexer demonstrate significant effectiveness when applied to LLaMA3-8B.

\noindent (2) {\bf RevMUX significantly enhances both performance and inference efficiency}:
Compared to Zero-Shot Prompting, RevMUX demonstrates a twofold increase in inference efficiency and improves accuracies by approximately $2\% - 10\%$ across the four datasets. Unlike the previous experiment, which established a strong baseline by training soft prompts for task domain adaptation, this study employs a manual prompt with a frozen LLaMA3 and demonstrates a clear overall performance gain brought by RevMUX. Our results suggest that during the training of reversible adapters, RevMUX also effectively learns to preserve the discriminative information that is helpful for classification tasks.

\subsection{Model Analysis and More Studies}
\label{sec:bi-result-model-analysis}

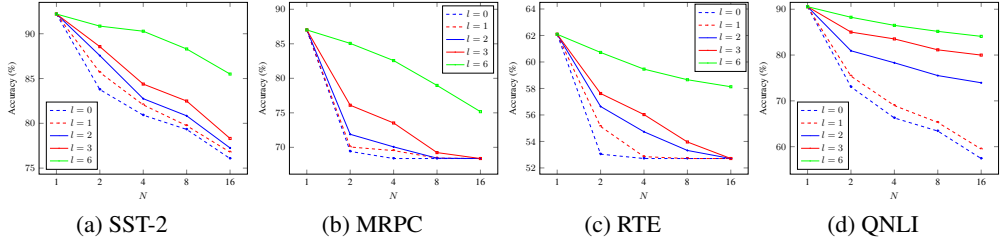
\begin{figure*}[!t]
  \centering
  \subfloat[SST-2]{
  \begin{tikzpicture}[scale=0.4]
  \pgfplotstableread{./res/sst-2.txt}\datatable
    \begin{axis}[
    font=\small,
    xlabel={$N$},
    xtick=data,
      xticklabels from table={\datatable}{[index] 0},
      xtick distance=1,
      table/x expr=\coordindex,
    ymax=93.50,
    ylabel={Accuracy (\%)},
    legend entries={$l=0$,$l=1$,$l=2$,$l=3$,$l=6$
    },
    mark size=1.0pt,
    legend pos= north east,
    legend cell align={left},
    legend style={font=\small,line width=.1pt,mark size=.1pt,
            at={(0.28,0.42)},
            /tikz/every even column/.append style={column sep=0.15em}},
    ],

  \addplot [blue,dashed,mark=*] table [x=data, y=L0] from \datatable;
  \addplot [red,dashed,mark=o] table [x=data, y=L1] from \datatable;

  \addplot [blue,thick,mark=+] table [x=data, y=L2] from \datatable;
  \addplot [red,thick,mark=square] table [x=data, y=L3] from \datatable;

  \addplot [green,thick,mark=square] table [x=data, y=L6] from \datatable;

  \end{axis}
  \end{tikzpicture}}
  \subfloat[MRPC]{
  \begin{tikzpicture}[scale=0.4]
  \pgfplotstableread{./res/mrpc.txt}\datatable
    \begin{axis}[
    font=\small,
    xlabel={$N$},
    xtick=data,
      xticklabels from table={\datatable}{[index] 0},
      xtick distance=1,
      table/x expr=\coordindex,
    ymax=91.00,
    ylabel={Accuracy (\%)},
    legend entries={$l=0$,$l=1$,$l=2$,$l=3$,$l=6$
    },
    mark size=1.0pt,
    legend pos= north east,
    legend cell align={left},
    legend style={font=\small,line width=.1pt,mark size=.1pt,
            at={(0.98,0.98)},
            /tikz/every even column/.append style={column sep=0.15em}},
    ],

  \addplot [blue,dashed,mark=*] table [x=data, y=L0] from \datatable;
  \addplot [red,dashed,mark=o] table [x=data, y=L1] from \datatable;

  \addplot [blue,thick,mark=+] table [x=data, y=L2] from \datatable;
  \addplot [red,thick,mark=square] table [x=data, y=L3] from \datatable;

  \addplot [green,thick,mark=square] table [x=data, y=L6] from \datatable;

  \end{axis}
  \end{tikzpicture}}
  \subfloat[RTE]{
  \begin{tikzpicture}[scale=0.4]
  \pgfplotstableread{./res/rte.txt}\datatable
    \begin{axis}[
    font=\small,
    xlabel={$N$},
    xtick=data,
      xticklabels from table={\datatable}{[index] 0},
      xtick distance=1,
      table/x expr=\coordindex,
    ymax=64.50,
    ylabel={Accuracy (\%)},
    legend entries={$l=0$,$l=1$,$l=2$,$l=3$,$l=6$
    },
    mark size=1.0pt,
    legend pos= north east,
    legend cell align={left},
    legend style={font=\small,line width=.1pt,mark size=.1pt,
            at={(0.99,0.99)},
            /tikz/every even column/.append style={column sep=0.15em}},
    ],

  \addplot [blue,dashed,mark=*] table [x=data, y=L0] from \datatable;
  \addplot [red,dashed,mark=o] table [x=data, y=L1] from \datatable;

  \addplot [blue,thick,mark=+] table [x=data, y=L2] from \datatable;
  \addplot [red,thick,mark=square] table [x=data, y=L3] from \datatable;

  \addplot [green,thick,mark=square] table [x=data, y=L6] from \datatable;

  \end{axis}
\end{tikzpicture}}
\subfloat[QNLI]{
\begin{tikzpicture}[scale=0.4]
\pgfplotstableread{./res/qnli.txt}\datatable
  \begin{axis}[
  font=\small,
  xlabel={$N$},
  xtick=data,
    xticklabels from table={\datatable}{[index] 0},
    xtick distance=1,
    table/x expr=\coordindex,
  ymax=91.50,
  ylabel={Accuracy (\%)},
  legend entries={$l=0$,$l=1$,$l=2$,$l=3$,$l=6$
  },
  mark size=1.0pt,
  legend pos= north east,
  legend cell align={left},
  legend style={font=\small,line width=.1pt,mark size=.1pt,
          at={(0.28,0.42)},
          /tikz/every even column/.append style={column sep=0.15em}},
  ],

\addplot [blue,dashed,mark=*] table [x=data, y=L0] from \datatable;
\addplot [red,dashed,mark=o] table [x=data, y=L1] from \datatable;

\addplot [blue,thick,mark=+] table [x=data, y=L2] from \datatable;
\addplot [red,thick,mark=square] table [x=data, y=L3] from \datatable;

\addplot [green,thick,mark=square] table [x=data, y=L6] from \datatable;

\end{axis}
\end{tikzpicture}}

\caption{Results of different $l$ and $N$ on $\mathrm{BERT}_{\mathrm{BASE}}$.}\label{fig:exp-l-N}
\end{figure*}

\begin{table*}[t!]
\setlength{\tabcolsep}{6pt}
    \centering\small
    \begin{tabular}{l| l c c c c c c }
    \toprule
    Backbone & Model & $N$ & SST-2 & MRPC & RTE & QNLI & Avg. Score \\
    \midrule
    \multirow{4}{*}{Llama3-8B} & Zero-Shot Prompting & 1
    & 92.64 & 70.10 & 72.92 & 76.99 & 78.16 \\
    \cline{2-8}
    ~ & Vanilla Adapters & 2
    & 94.01 & 80.96 & 82.72 & 85.99 & 85.92 \\
    ~ & Only Multiplexer Reversible & 2
    & 94.09 & 81.08 & 82.82 & 86.24 & 86.06 \\
    ~ & RevMUX & 2
    & {\bf 94.38} & {\bf 81.30} & {\bf 83.18} & {\bf 86.53} & {\bf 86.35} \\
    \bottomrule
\end{tabular}
\caption{Llama3-8B results on the four datasets of GLUE benchmark.\label{table:bi-exp-glue-llama}}
\end{table*}

\begin{table}[t!]\small
    \centering
    \tabcolsep 6.0pt
    \begin{tabular}{c| c c c c c}
    \toprule
    $l$ & 0 & 1 & 2 & 3 & 6 \\
    \midrule
    $\nearrow$ & 207\% & 198\% & 189\% & 181\% & 154\% \\
    \bottomrule
\end{tabular}
    \caption{Inference efficiency improvement with different prefilling layers on SST-2 with $\mathrm{BERT}_{\mathrm{BASE}}$.}\label{table:bi-experiment-l-efficiency}
\end{table}

We analyze the performance and inference efficiency of RevMUX (\FE) by varying the number of prefilling layers $l$, the batch size $N$, and the impact of InfoNCE loss $\lambda$. We use ${\mathrm{BERT}}_{\mathrm{BASE}}$ and report accuracy on MRPC and RTE in Figure~\ref{fig:exp-l-N} and speedups ($\nearrow$) on SST-2 in Table \ref{table:bi-experiment-l-efficiency}.

\noindent {\bf The impact of $N$ and $l$ on performance}: Figure~\ref{fig:exp-l-N} shows a clear downward trend in classification accuracy as $N$ increases. This is anticipated, as mixing more samples in a batch makes it more difficult for RevMUX to preserve the individual distinctiveness given the limited capacity of the reversible modules, $\mathcal{F}$ and $\mathcal{G}$. However, with a sufficient number of prefilling layers (e.g., $l=6$), the model can maintain relatively high accuracy even when $N$ is increased to 16. This suggests that increasing the number of prefilling layers is an effective strategy to enhance model performance, allowing it to sustain accuracy despite larger $N$ values. More studies can be found in Table~\ref{table:bi-experiment-bert-glue-n} in the Appendix.

\noindent {\bf The impact of prefilling on efficiency}: While Figure \ref{fig:exp-l-N} indicates that increasing the number of prefilling layers enhances classification accuracy, it also raises a concern about inference efficiency. As shown in Table~\ref{table:bi-experiment-l-efficiency}, increasing the number of prefilling layers to 6 can reduce the speedup by $50\%$ compared to not using any prefilling. However, as higher layers in LLMs typically provide a better representation space that may help in distinguishing different samples, choosing the optimal number of prefilling layers remains a trade-off to balance accuracy and efficiency.

\section{Conclusion}
In this paper, we introduce RevMUX, a parameter-efficient data multiplexing framework designed to enhance the batch inference efficiency of LLMs. RevMUX features a reversible multiplexer that combines multiple samples, allowing the demultiplexer to reverse this process and restore individual outputs for classification. We train RevMUX on downstream tasks while keeping the backbone LLM frozen, and apply it for batch inference. Extensive experiments on BERT-base, T5 across three scales, and LLaMA3-8B demonstrate the effectiveness of RevMUX in enhancing both accuracy and efficiency. Ablation studies confirm the synergistic function of the reversible multiplexer and the reverse demultiplexer.

\section*{Acknowledgements}

This research is supported, in part, by the Joint NTU-WeBank Research Centre on Fintech (Award No. NWJ-2020-007), Nanyang Technological University, Singapore. This research is also supported, in part, by the National Research Foundation, Prime Minister’s Office, Singapore under its NRF Investigatorship Programme (NRFI Award No. NRF-NRFI05-2019-0002). Xu Guo thanks Wallenberg-NTU Presidential Postdoctoral Fellowship. Zhiwei Zeng thanks the support from the Gopalakrishnan-NTU Presidential Postdoctoral Fellowship. Any opinions, findings and conclusions or recommendations expressed in this material are those of the authors and do not reflect the views of National Research Foundation, Singapore.

\newpage

\section*{Limitations}
While RevMUX presents a promising step forward in improving LLM inference efficiency, several limitations need to be acknowledged.

\noindent {\bf Inference efficiency-performance trade-offs}: Although RevMUX effectively improves inference efficiency, there is an inherent trade-off with potential loss in inference performance. While our experiments show that RevMUX can largely retain a satisfactory classification performance in the majority of scenarios, the performance compromises could vary with different datasets or tasks. For instance, we observe that the efficiency-performance trade-off is more pronounced on the RTE dataset with $\mathrm{T5}_{\mathrm{Large}}$ and $\mathrm{T5}_{\mathrm{Base}}$. This may be attributed to the inherent complexity and nuances of the RTE dataset, and the underlying causes warrant further investigation.

\noindent {\bf Potential for bias and fairness issues}: As with many other AI and ML methods, there is a risk that the multiplexing strategy could inadvertently amplify existing biases in the data. Proper handling of fairness and bias relation issues in data multiplexing remains an area requiring further investigation.

\noindent {\bf Further empirical evidence on scalability}: While RevMUX shows promise in enhancing efficiency, its scalability for extremely large-scale deployments or real-time applications needs thorough evaluation. Our experimental results suggest that larger backbones tend to experience greater performance compromises to gain efficiency. Understanding how RevMUX scales with even larger model sizes and deployment contexts is critical for broader applications.

\section*{Ethics Statement}
In conducting this research, we have adhered to ethical standards and have not introduced any new ethical concerns:
\begin{itemize}
  \item \textbf{Data usage}: We did not release any new datasets as part of this study. All datasets used are publicly available or have been licensed for academic purposes. We ensure compliance with the data usage policies of these sources.
  \item \textbf{Codes and artefacts}: The source code for baselines and other artefacts employed in our study are either open-sourced or licensed for academic use.
  \item \textbf{Transparency and accountability}: We strive for transparency in our research. All results and methodologies are clearly documented, and we encourage replication and scrutiny by the research community.
\end{itemize}



\bibliography{nlp}
\bibliographystyle{acl_natbib}

\input{appendix}

\end{document}

%% file: appendix.tex
\appendix
\section*{Appendix}

\section{RevMUX ($N>2$)}
\label{sec:bi-appendix-n-gt-2}

Similar with the RevMUX pipeline when $N=2$, the pipeline of RevMUX ($N>2$) has a slightly different multiplexer and demultiplexer to adapt the condition of $N$.

\subsection{Reversible Multiplexer}
In order to keep the reversible design, when $N>2$, we can expand the Eq~\eqref{eq:bi-methodology-ig} as:
\begin{align}
  \label{eq:bi-methodology-ig-n}
  \bo_1^l&=\bi_1^l+\cF_1(\bi_N^l),\\\nonumber
  \bo_2^l&=\bi_2^l+\cF_2(\bo_1^l),\\\nonumber
  \bo_3^l&=\bi_3^l+\cF_3(\bo_2^l),\\\nonumber
  &\cdots\\\nonumber
  \bo_N^l&=\bi_N^l+\cF_N(\bo_{N-1}^l),\\\nonumber
  \bo^l&=\mathrm{concat}[\bo_1^l,\bo_2^l,\bo_3^l,\cdots,\bo_N^l].
\end{align}
It is notably that $\cF(\cdot)$ and $\cG(\cdot)$ in Eq.~\eqref{eq:bi-methodology-ig} is the same as $\cF_1(\cdot)$ and $\cF_2(\cdot)$ in Eq~\eqref{eq:bi-methodology-ig-n}.

\subsection{Reverse Demultiplexer}
Similar with Eq~\eqref{eq:bi-methodology-ig-n}, we expand the Eq~\eqref{eq:bi-methodology-dg} when $N>2$ as:
\begin{align}
  \label{eq:bi-methodology-dg-n}
  [\hat{\bo}_1,\hat{\bo}_2,\hat{\bo}_3,\cdots,\hat{\bo}_N]&=\hat{\bo},\\\nonumber
  \hat{\bi}_N&=\hat{\bo}_N-\cF_N(\hat{\bo}_{N-1}),\\\nonumber
  \hat{\bi}_{N-1}&=\hat{\bo}_{N-1}-\cF_{N-1}(\hat{\bo}_{N-2}),\\\nonumber
  &\cdots\\\nonumber
  \hat{\bi}_1&=\hat{\bo}_1-\cF_1(\hat{\bi}_N).
\end{align}
In summary, Eq~\eqref{eq:bi-methodology-ig} and Eq~\eqref{eq:bi-methodology-dg} are the special case ($N=2$) of Eq~\eqref{eq:bi-methodology-ig-n} and Eq~\eqref{eq:bi-methodology-dg-n}, respectively.

\section{Datasets}

The detailed statistics of the datasets is shown in Table~\ref{table:bi-dataset-stat}.
We used four datasets from the GLUE benchmark to evaluate our models. The SST-2 dataset, with 67,349 training samples and 872 evaluation samples, is used for binary sentiment classification, labelling sentences as either positive or negative. The MRPC dataset consists of 3,668 training samples and 408 evaluation samples, involving sentence pairs labelled to indicate whether they are paraphrases of each other. The QNLI dataset includes 104,743 training samples and 5,463 evaluation samples, where the task is to determine if a given sentence correctly answers a question, derived from the Stanford Question Answering Dataset (SQuAD). Lastly, the RTE dataset, with 2,490 training samples and 277 evaluation samples, involves binary classification to determine whether one sentence entails another.

\begin{table}[t!]\small
    \centering
    \tabcolsep 2.0pt
    \begin{tabular}{c| c c c}
    \toprule
    Dataset & \# Labels & \# Train samples & \# Evaluation samples \\
    \midrule
    SST-2 & 2 & 67,349 & 872 \\
    MRPC & 2 & 3,668 & 408 \\
    QNLI & 2 & 104,743 & 5,463 \\
    RTE & 2 & 2,490 & 277 \\
    \bottomrule
\end{tabular}
    \caption{Summary statistics of four datasets from GLUE benchmark. We evaluate all models on the development set of all datasets.
    }\label{table:bi-dataset-stat}
\end{table}

\section{Performance Testing}
\label{sec:bi-appendix-evaluation}
\subsection{Testing Rounds}
\label{sec:bi-appendix-testing-rounds}

It is important to note that RevMUX does not adhere to the commutative property. For instance, $\mathrm{RevMUX}(x_1, x_2)$ is not necessarily equal to $\mathrm{RevMUX}(x_2, x_1)$. Unlike conventional minibatch processing pipelines that eliminate randomness during inference, RevMUX is sensitive to the order of inputs. As a result, the same input instance can yield different predictions depending on the testing order. Therefore, to achieve a more robust and accurate evaluation, it is essential to assess RevMUX across multiple rounds, with varying input sequences (e.g., $[x_1, x_2, x_3, x_4]$ versus $[x_1, x_4, x_3, x_2]$).

\begin{figure*}
  \centering
  \includegraphics[width=\textwidth]{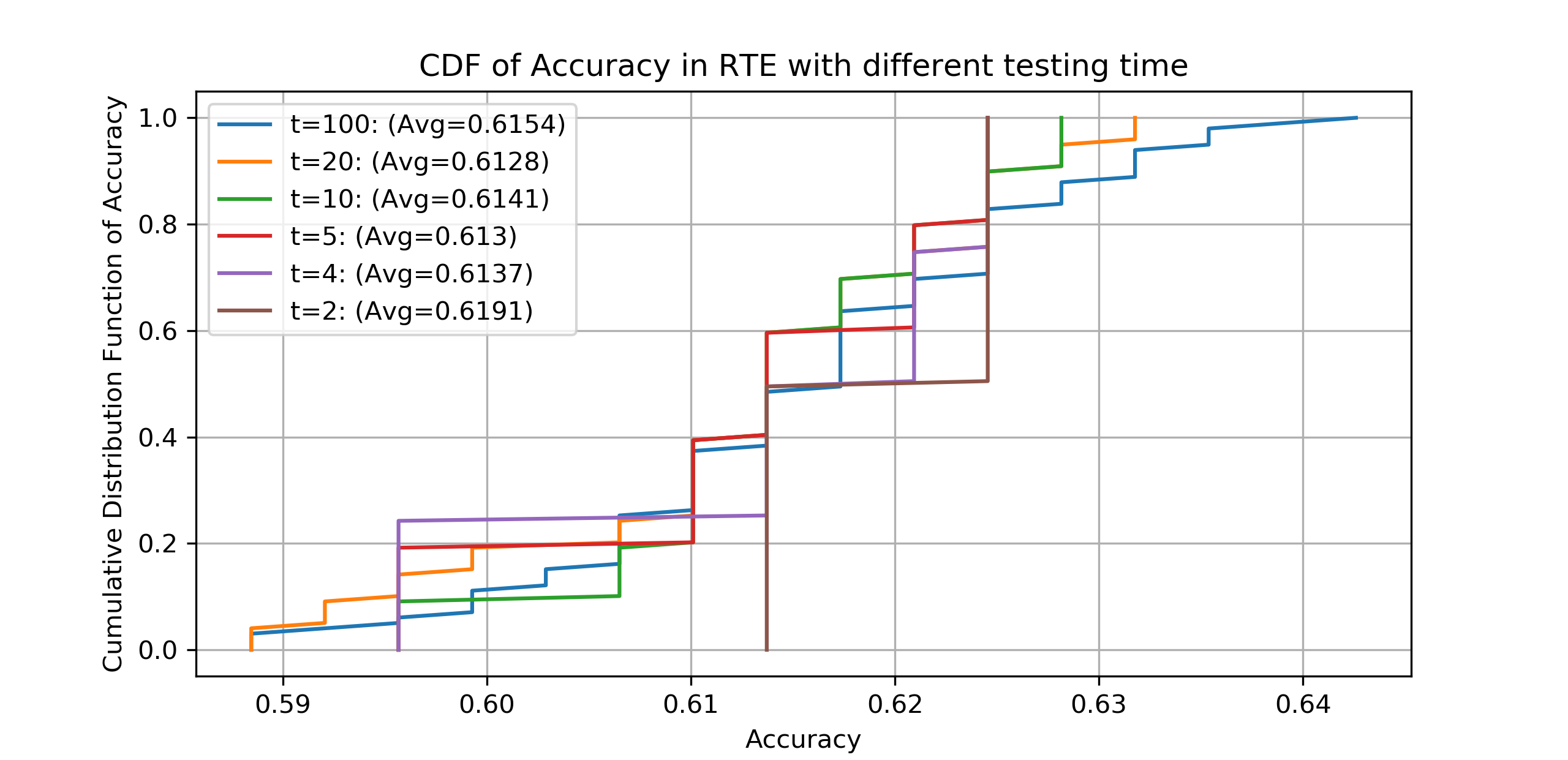}

  \caption{CDF of testing times $t$.}
  \label{fig:bi-exp-cdf-of-t}
\end{figure*}

To empirically explore the appropriate number of testing rounds, we fixed a RevMUX configuration and evaluated the model across multiple rounds, recording the cumulative distribution function (CDF) of accuracy. As illustrated in Figure~\ref{fig:bi-exp-cdf-of-t}, we observe that as the number of testing rounds, $t$, increases, the distribution of the model evaluation accuracy becomes smoother. Our analysis suggests that $t=100$ provides a sufficiently robust evaluation. However, it is notable that the CDF curve for $t=10$ closely approximates that of $t=100$. Therefore, we selected $t=10$ for our evaluations to achieve a balance between efficiency and accuracy.

\subsection{The Effect of Fine-tuning on Performance during Data Multiplexing}

To assess the impact of fine-tuning versus not fine-tuning the $\mathrm{BERT}_{\mathrm{BASE}}$ backbone on the performance, we conducted a $t$-test to evaluate the significance of fine-tuning (RevMUX \FT) versus not fine-tuning (RevMUX \FE). As shown in Table~\ref{table:bi-experiment-p-value}, the results indicate significant performance improvements with fine-tuning the backbone model across all datasets. For SST-2 and MRPC, the p-values ($0.011$ and $<0.005$, respectively) and negative t-statistics ($-2.833$ and $-3.712$) demonstrate that fine-tuning yields superior accuracy. The RTE dataset shows an exceptionally high t-statistic of $-73.688$ with a p-value <0.0001, highlighting a dramatic performance boost with fine-tuning. Similarly, for QNLI, the strong negative t-statistic of -5.603 and a p-value <0.0001 confirm the advantages of fine-tuning.

\begin{table}[t!]\small
    \centering
    \tabcolsep 6.0pt
    \begin{tabular}{c| c c c c c}
    \toprule
    Dataset & SST-2 & MRPC & RTE & QNLI \\
    \midrule
    $T$-statistic & $-2.833$ & $-3.712$ & $-73.688$ & $-5.603$ \\
    $p$-value & $0.011$ & $<0.005$ & $<0.0001$ & $<0.0001$ \\
    \bottomrule
\end{tabular}
    \caption{T-statistic and the $p$-value of RevMUX (\FT) outperforms RevMUX (\FE).}\label{table:bi-experiment-p-value}
\end{table}

\section{Scalability Test}
\label{sec:bi-appendix-scaling-llama}
\begin{table*}[t!]\small
    \centering
    \tabcolsep 6.0pt
    \begin{tabular}{c| p{0.7\textwidth}}
    \toprule
    Dataset & Template \\
    \midrule
    \multirow{3}*{SST-2} & {\bf User}: \texttt{You are require to predict the sentiment (positive or negative) to the following sentence. You should response positive or negative, only one token is accepted.} \\
    ~ & {\bf User}: \texttt{<|start of the sentence|>:} <sentence> \texttt{<|end of the sentence|>.}\\
    ~ & {\bf Assistant}: {\bf ?}\\
    \midrule
    \multirow{4}*{RTE, MRPC, QNLI} & {\bf User}: \texttt{You are require to predict the two following sentences are entailment or not (yes or no). You should response yes or no, only one token is accepted.}\\
    ~ & {\bf User}: \texttt{<|start of the sentence1|>:} <sentence1> \texttt{<|end of the sentence1|>}\\
    ~ & {\bf User}: \texttt{<|start of the sentence2|>:} <sentence2> \texttt{<|end of the sentence2|>}\\
    ~ & {\bf Assistant}: {\bf ?}\\
    \midrule
\end{tabular}
    \caption{Chat template for LLaMA3-8B-Instruct. Here ``<sentence>'' indicates single-sentence classification, ``<sentence1>'' and ``<sentence2>'' indicate the pair-wised sentence classification.}\label{table:bi-experiment-llama-template}
\end{table*}

\subsection{Scaling to Larger Model Size}

\subsubsection{Backbone Selection}

To evaluate the effectiveness of RevMUX on larger backbone models, we selected the recently released and well-known open-source LLM, LLaMA3. Due to limited computational resources, we opted for the 8B model variant, which can be trained on a V100 GPU with 32GB of memory. To maintain consistency with the pre-training scenarios, we employed a chat template. Based on these considerations, we selected LLaMA3-8B-Instruct as the backbone for our experiments.

\subsubsection{Implemtation Details}

Given that SST-2 is a single-sentence classification task, while RTE, MRPC, and QNLI are pairwise sentence classification tasks, we utilized two different chat templates, as illustrated in Table~\ref{table:bi-experiment-llama-template}. To simplify the verbalizer for answer prediction, we imposed a constraint that   ``only one token is accepted'' and selected the language head prediction of the final token as the prediction for LLaMA. By applying this chat template for zero-shot transfer, the results presented in Table~\ref{table:bi-exp-glue-llama} validate the effectiveness of our approach.

\subsection{Scaling to Larger $N$}
\label{sec:bi-appendix-model-analysis-N}

To explore the scalability of RevMUX with varying values of $N$, we conduct a comparative experiment against $\mathrm{MUX}$-$\mathrm{PLM}$ using the ${\mathrm{BERT}}_{\mathrm{BASE}}$ backbone. The results, presented in Table~\ref{table:bi-experiment-bert-glue-n}, lead to the following key observations:

\noindent (1) {\bf RevMUX outperforms $\mathrm{MUX}$-$\mathrm{PLM}$ when $N=2$}: Under a fair comparison, RevMUX achieves an average score of 81.22 when $N=2$, surpassing the 80.19 score of $\mathrm{MUX}$-$\mathrm{PLM}$.

\noindent (2) {\bf RevMUX maintains comparable or superior performance with larger $N$ values}: Notably, as $N$ increases, RevMUX continues to demonstrate its scalability. For instance, the average score of RevMUX with $N=8$ (77.78) is comparable to that of $\mathrm{MUX}$-$\mathrm{PLM}$ with $N=5$ (77.92). Furthermore, RevMUX with $N=16$ achieves a higher average score (75.72) than $\mathrm{MUX}$-$\mathrm{PLM}$ with $N=10$ (75.61), highlighting the effectiveness and scalability potential of RevMUX.

\begin{table}[t!]\small
    \centering
    \tabcolsep 3.0pt
    \begin{tabular}{l| c c c c c}
    \toprule
    ~ & SST-2 & MRPC & RTE & QNLI & Avg. Score \\
    \midrule
    With InfoNCE & {\bf 89.14} & {\bf 82.45} & {\bf 60.22} & {\bf 85.63} & {\bf 79.36} \\
    w.o. InfoNCE & 89.03 & 82.11 & 58.45 & 85.40 & 78.75 \\
    \bottomrule
\end{tabular}
    \caption{Ablation study results on with vs without InfoNCE loss on $\mathrm{T5}_{\mathrm{Small}}$.}\label{table:bi-experiment-ablation-infonce-t5}
\end{table}

\begin{table}[t!]\small
    \centering
    \tabcolsep 3.0pt
    \begin{tabular}{l| c c c c c}
    \toprule
    ~ & SST-2 & MRPC & RTE & QNLI & Avg. Score \\
    \midrule
    With InfoNCE & {\bf 90.85} & {\bf 85.06} & {\bf 60.72} & {\bf 88.25} & {\bf 81.22} \\
    w.o. InfoNCE & 90.58 & 84.04 & 58.59 & 87.85 & 80.27 \\
    \bottomrule
\end{tabular}
    \caption{Ablation results about with vs without InfoNCE loss on $\mathrm{BERT}_{\mathrm{BASE}}$.}\label{table:bi-experiment-ablation-infonce}
\end{table}

\begin{table*}[t!]
    \centering
    \tabcolsep 6.0pt
    \begin{tabular}{l| c c c c c c c}
    \toprule
    Model & $N$ & Tuned & SST-2 & MRPC & RTE & QNLI & Avg. Score \\
    \midrule
    $\mathrm{MUX}$-${\mathrm{PLM}}$ & 1 & \FT & 91.74 & 87.75 & 63.18 & 90.54 & 83.30 \\
    \midrule
    RevMUX & 2 & \FE & 90.85 & 85.06 & 60.72 & 88.25 & 81.22 \\
    $\mathrm{MUX}$-${\mathrm{PLM}}$ & 2 & \FT & 90.62 & 83.77 & 58.19 & 88.17 & 80.19 \\
    \midrule
    RevMUX & 4 & \FE & 90.28 & 82.57 & 59.46 & 86.48 & 79.70 \\
    $\mathrm{MUX}$-${\mathrm{PLM}}$ & 5 & \FT & 86.88 & 80.10 & 59.13 & 85.58 & 77.92 \\
    RevMUX & 8 & \FE & 88.30 & 78.97 & 58.66 & 85.17 & 77.78 \\
    $\mathrm{MUX}$-${\mathrm{PLM}}$ & 10 & \FT & 83.44 & 78.63 & 58.27 & 82.08 & 75.61 \\
    RevMUX & 16 & \FE & 85.50 & 75.17 & 58.13 & 84.08 & 75.72 \\
    \bottomrule
\end{tabular}
    \caption{Model comparison of RevMUX and $\mathrm{MUX}$-${ \mathrm{PLM}}$~\cite{DBLP:conf/emnlp/MurahariDJSW0N23} using $\mathrm{BERT}_{\mathrm{BASE}}$ as backbone model with different $N$.}\label{table:bi-experiment-bert-glue-n}
\end{table*}

\section{Hyperparameter Analysis}
\subsection{Impacts of $\lambda$ for InfoNCE Loss}
\label{sec:bi-appendix-model-analysis-lambda}

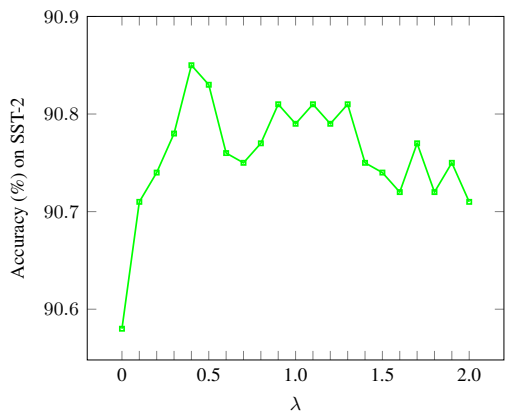
\begin{figure}[!h]
    \centering
    \begin{tikzpicture}[scale=0.8]
    \pgfplotstableread{./res/lambda1.txt}\datatable
      \begin{axis}[
      font=\small,
      xlabel={$\lambda$},
      xtick=data,
      xticklabels={0,,,,,0.5,,,,,1.0,,,,,1.5,,,,,2.0,},
        xtick distance=1,
      ymax=90.90,
      ylabel={Accuracy (\%) on SST-2},
      mark size=1.0pt,
      legend pos= north east,
      legend cell align={left},
      legend style={font=\small,line width=.1pt,mark size=.1pt,
              at={(0.99,0.99)},
              /tikz/every even column/.append style={column sep=0.15em}},
      ],

    \addplot [green,thick,mark=square] table [x=lambda, y=Accuracy] from \datatable;

    \end{axis}
    \end{tikzpicture}
    \caption{The impact of different $\lambda$ for InfoNCE loss under the $\mathrm{BERT}_{\mathrm{BASE}}$ backbone.}
    \label{fig:bi-experiment-lambda}
\end{figure}

\noindent {\bf Impacts of InfoNCE Loss}: In order to explore the effectiveness of InfoNCE in our framework, we conduct ablation studies about with and without InfoNCE Loss with $\mathrm{BERT}_{\mathrm{BASE}}$ backbone. As shown in Table~\ref{table:bi-experiment-ablation-infonce}, the InfoNCE loss improves the average score from 80.27 to 81.22, demonstrates the effectiveness of the objective. More detailed analysis of the InfoNCE loss can be found in Appendix~\ref{sec:bi-appendix-model-analysis-lambda}.

In this section, we extend our experiments to further investigate the impact of the InfoNCE loss.

As shown in Table~\ref{table:bi-experiment-ablation-infonce-t5}, we observe that incorporating the InfoNCE loss leads to improvements across all four datasets using the $\mathrm{T5}_{\mathrm{Small}}$ backbone. This aligns with the findings from the $\mathrm{BERT}_{\mathrm{BASE}}$ backbone discussed in Section~\ref{sec:bi-result-model-analysis}, demonstrating the consistent effectiveness of the InfoNCE loss.

To gain deeper insights, we also conduct experiments varying the value of $\lambda$ in Eq~\eqref{eq:bi-loss-lambda}. As illustrated in Figure~\ref{fig:bi-experiment-lambda}, we find that a value around 0.5 yields the best performance, and we adopt this setting for the subsequent experiments.

\section{Inference Efficiency Comparison}

To compare inference efficiency, we report the FLOPs required for validation set inference. For a fair comparison, we set the batch size to 32 and the sequence length to 128, following the methodology of~\cite{DBLP:conf/emnlp/MurahariDJSW0N23}. The efficiency improvement, denoted in column $\nearrow$, is calculated based on the average FLOPs used across all four datasets. The results are presented in Table~\ref{table:bi-efficiency-comparison-bert} and Table~\ref{table:bi-efficiency-comparison-t5}.

Based on the inference efficiency results presented in Table~\ref{table:bi-efficiency-comparison-bert}, we evaluated various models using ${\mathrm{BERT}}_{\mathrm{BASE}}$ as the backbone. RevMUX (\FE), despite achieving comparable efficiency, shows slightly higher average FLOPs (33.713 T) compared to $\mathrm{DataMUX}$ (28.799 T) and $\mathrm{MUX}$-${ \mathrm{BERT}}_{\mathrm{BASE}}$ (25.834 T).

Based on the inference efficiency results presented in Table~\ref{table:bi-efficiency-comparison-t5}, using T5 as the backbone model, RevMUX achieves about $45\%$ speedups across all scales. RevMUX shows average FLOPs of 8.188 T, 34.532 T, and 119.588 T on $\mathrm{T5}_{\mathrm{Small}}$, $\mathrm{T5}_{\mathrm{Base}}$, and $\mathrm{T5}_{\mathrm{Large}}$, respectively. The speed-up percentages on different $\mathrm{T5}$ backbones are roughly around 140\%, ranging from 138\% to 144\%.

\begin{table*}[t!]
\setlength{\tabcolsep}{2pt}
    \centering\small
    \begin{tabular}{c l| c c c c c c c c c c}
    \toprule
    ~ & Model & $N$ & $\nearrow$ & Tuned & SST-2 & MRPC & RTE & QNLI & Avg. FLOPs \\
    \midrule
    Backbones & $\mathrm{MUX}$-${ \mathrm{BERT}}_{\mathrm{BASE}}$~\cite{DBLP:conf/emnlp/MurahariDJSW0N23} & 1 & 100\% & \FT & 25.824 & 11.477 & 7.651 & 162.593 & 51.886 \\
    \midrule
    \midrule
    \multirow{2}{*}{Baselines} & $\mathrm{DataMUX}$~\cite{DBLP:conf/nips/MurahariJYN22} & 2 & 180\% & \FT & 13.866 & 6.400 & 4.267 & 90.664 & 28.799 \\
    ~ & $\mathrm{MUX}$-${ \mathrm{BERT}}_{\mathrm{BASE}}$~\cite{DBLP:conf/emnlp/MurahariDJSW0N23} & 2 & 201\% & \FT & 12.439 & 5.741 & 3.827 & 81.330 & 25.834 \\
    \midrule
    \multirow{4}{*}{Ours} & Vanilla Adapters & 2 & 156\% & \FE & 16.545 & 7.741 & 5.263 & 103.663 & 33.303  \\
    ~ & Only Multiplexer Reversible & 2 & 161\% & \FE & 16.019 & 7.495 & 5.096 & 100.363 & 32.243 \\
    ~ & RevMUX & 2 & 154\% & \FE & 16.749 & 7.837 & 5.328 & 104.938 & 33.713 \\
    \bottomrule
\end{tabular}
\caption{Inference efficiency comparison using $\mathrm{BERT}_{\mathrm{BASE}}$ as backbone model. (Unit: T FLOPs)
\label{table:bi-efficiency-comparison-bert}}
\end{table*}

\begin{table*}[t!]
\setlength{\tabcolsep}{6pt}
    \centering\small
    \begin{tabular}{l| l c c c c c c c c c c}
    \toprule
    Backbone & Model & $N$ & Tuned & $\nearrow$ & SST-2 & MRPC & RTE & QNLI & Avg. FLOPs \\
    \midrule
    \multirow{4}{*}{$\mathrm{T5}_{\mathrm{Small}}$} & Task-specific Backbone & 1 & \FT & 100\%
    & 5.919 & 2.770 & 1.880 & 37.084 & 11.913 \\
    \cline{2-10}
    ~ & Vanilla Adapters & 2 & \FE & 138\%
    & 4.293 & 2.008 & 1.366 & 26.891 & 8.640 \\
    ~ & Only Multiplexer Reversible & 2 & \FE & 146\%
    & 4.058 & 1.899 & 1.291 & 25.424 & 8.168 \\
    ~ & RevMUX & 2 & \FE & 145\%
    & 4.068 & 1.903 & 1.294 & 25.487 & 8.188 \\
    \midrule
    \midrule
    \multirow{4}{*}{$\mathrm{T5}_{\mathrm{Base}}$} & Task-specific Backbone & 1 & \FT & 100\%
    & 24.689 & 11.552 & 7.843 & 154.677 & 49.690 \\
    \cline{2-10}
    ~ & Vanilla Adapters & 2 & \FE & 140\%
    & 17.660 & 8.263 & 5.618 & 110.644 & 35.546 \\
    ~ & Only Multiplexer Reversible & 2 & \FE & 144\%
    & 17.133 & 8.016 & 5.451 & 107.344 & 34.486 \\
    ~ & RevMUX & 2 & \FE & 144\%
    & 17.156 & 8.027 & 5.458 & 107.486 & 34.532 \\
    \midrule
    \midrule
    \multirow{4}{*}{$\mathrm{T5}_{\mathrm{Large}}$} & Task-specific Backbone & 1 & \FT & 100\%
    & 84.782 & 39.668 & 26.932 & 531.149 & 170.633 \\
    \cline{2-10}
    ~ & Vanilla Adapters & 2 & \FE & 141\%
    & 60.308 & 28.218 & 19.187 & 377.854 & 121.392 \\
    ~ & Only Multiplexer Reversible & 2 & \FE & 143\%
    & 59.372 & 27.777 & 18.888 & 371.987 & 119.506 \\
    ~ & RevMUX & 2 & \FE & 143\%
    & 59.412 & 27.798 & 18.901 & 372.239 & 119.588 \\
    \bottomrule
\end{tabular}
\caption{Inference efficiency comparison using T5 as backbone model. (Unit: T FLOPs)\label{table:bi-efficiency-comparison-t5}}
\end{table*}